\documentclass[letterpaper, 10 pt, journal, twoside]{IEEEtran}
\usepackage{amsmath,amsfonts}
\usepackage{array}
\usepackage[caption=false,font=scriptsize]{subfig} 
\usepackage{textcomp}
\usepackage{stfloats}
\usepackage{url}
\usepackage{verbatim}
\usepackage{graphicx}
\usepackage{cite}
\usepackage{hyperref}
\usepackage{bm}
\usepackage[ruled,linesnumbered]{algorithm2e}
\usepackage{amssymb}  

\DeclareMathOperator*{\argminA}{arg\,min} 
\usepackage{adjustbox}

\begin{document}

\title{Visual Servoing NMPC Applied to UAVs \\for Photovoltaic Array Inspection}

\author{Edison Velasco-Sánchez\IEEEauthorrefmark{1}\IEEEauthorrefmark{4}, Luis F. Recalde\IEEEauthorrefmark{2}\IEEEauthorrefmark{3}\IEEEauthorrefmark{4}, Bryan S. Guevara\IEEEauthorrefmark{2}\IEEEauthorrefmark{4}, José Varela-Aldás\IEEEauthorrefmark{3}, \\ Francisco A. Candelas\IEEEauthorrefmark{1}, Santiago T. Puente\IEEEauthorrefmark{1} and Daniel C. Gandolfo\IEEEauthorrefmark{2}
\thanks{Manuscript received: October 01, 2023; Revised: November 27, 2023; Accepted: January 13, 2024.}
\thanks{This paper was recommended for publication by Editor Giuseppe Loianno upon evaluation of the Associate Editor and Reviewers’ comments.}
\thanks{This work has been supported by the Ministry of Science and Innovation of the Spanish Government through the research project PID2021-122685OBI00 and the grant PRE2019-088069 for the training of research PhD staff.}

\thanks{\IEEEauthorrefmark{4} These authors contributed equally.}
\thanks{\IEEEauthorrefmark{1} Edison Velasco-Sánchez, Francisco A. Candelas, and Santiago T. Puente are with the Group of Automation, Robotics and Computer Vision (AUROVA), University of Alicante, San Vicente del Raspeig S/N, Alicante, Spain.
        {\tt\small edison.velasco@ua.es}}
\thanks{\IEEEauthorrefmark{2} Luis F. Recalde, Bryan S. Guevara and Daniel C. Gandolfo are with Instituto de Automática (INAUT), Universidad Nacional de San Juan-CONICET, Argentina.
        {\tt\small lrecalde@inaut.unsj.edu.ar,
            bguevara@inaut.unsj.edu.ar}}
\thanks{\IEEEauthorrefmark{3} Luis F. Recalde and José Varela-Aldás are with Centro de Investigaciones de Ciencias Humanas y de la Educación (CICHE), Universidad Tecnológica Indoamérica, Ambato, Ecuador.
        {\tt\small fernandorecalde@uti.edu.ec, josevarela@uti.edu.ec}}
\thanks{Digital Object Identifier (DOI):  10.1109/LRA.2024.3360876.}
}
\markboth{IEEE ROBOTICS AND AUTOMATION LETTERS. PREPRINT VERSION. ACCEPTED January, 2024}
{Velasco-Sánchez \MakeLowercase{\textit{et al.}}: Visual Servoing NMPC Applied to UAVs for Photovoltaic Array Inspection}

\maketitle
\begin{abstract}
The photovoltaic (PV) industry is seeing a significant shift toward large-scale solar plants, where traditional inspection methods have proven to be time-consuming and costly. Currently, the predominant approach to PV inspection using unmanned aerial vehicles (UAVs) is based on the capture and detailed analysis of aerial images (photogrammetry). However, the photogrammetry approach presents limitations, such as an increased amount of useless data and potential issues related to image resolution that negatively impact the detection process during high-altitude flights.
In this work, we develop a visual servoing control system with dynamic compensation using nonlinear model predictive control (NMPC) applied to a UAV. This system is capable of accurately tracking the middle of the underlying PV array at various frontal velocities and height constraints, ensuring the acquisition of detailed images during low-altitude flights. The visual servoing controller is based on extracting features using \hbox{RGB-D} images and employing a Kalman filter to estimate the edges of the PV arrays. Furthermore, this work demonstrates the proposal in both simulated and real-world environments using the commercial aerial vehicle (DJI Matrice 100), with the purpose of showcasing the results of the architecture. Our approach is available to the scientific community in: \href{https://github.com/EPVelasco/VisualServoing_NMPC}{https://github.com/EPVelasco/VisualServoing\_NMPC.}
\end{abstract}


%%%%%%%%%%%%%%%%%%%%%%%%%%%%%%%%%%%%%%%%%%%%%%%%%%%%%%%%%%%%%%%%%%%%%%%%%

\section{Introduction}
\label{sec:Introduction}
\IEEEPARstart{N}{owadays}, society is facing global energy challenges and the search for alternative sources to fossil fuels \cite{Heinberg2016OurEnergy}. Renewable energies (RE), such as wind, hydro and solar, are abundant and are not at risk of depletion \cite{Bhuiyan2022RenewableReview, Gnangoin2022RenewableCapital}.
The PV sector has grown rapidly and has become the primary source of renewable energy, given the abundance and accessibility of solar power \cite{Sovacool2022TowardsPanels, Tverberg2012OilCrisis}. However, PV installations are limited by changing environments, where elements such as sand, wind, and dust significantly reduce the efficiency of PV modules \cite{Bessa2021MonitoringStrategies}. In addition, this type of system requires periodic maintenance due to the need for optimal conditions to function properly \cite{Golnas2013PVPerspective}. Many studies have shown that a large number of PV modules present maintenance-related damage \cite{Grimaccia2017SurveyItaly}, which can even reach $80\%$ of the total installed modules, where the most common problem is strictly related to surface defects such as mismatches, cracks, discolorations, snail trails, and soiling \cite{Djordjevic2014DetectableAustralia,Pedersen2016EffectNorway,Aghaei2022ReviewModules,Tsanakas2016FaultsChallenges}.
 
In recent years, PV modules have been inspected using different techniques. One method is \textit{Electroluminescence-based inspection}, involving the application of direct current to the module of PV array and the measurement of the resulting photoemission \cite{Pratt2021DefectSegmentation}. Another approach is \textit{I-V measurements}, where I-V curves are generated using information from PV modules, but the exact location of the failure is not provided \cite{WaqarAkram2022FailuresReview}. \textit{Infrared thermography} utilizes thermal cameras to capture radiation and convert it into thermogram images \cite{Tang2016QuantitativeTechnology}. Furthermore, \textit{ Visual inspection} by experts offers a simple method for identifying potential defects in PV modules \cite{Djordjevic2014DetectableAustralia}. However, visual inspection can be time-consuming, making it more suitable for small plants; on a large scale, it is almost impractical.

\begin{figure*}[!b]
    \centering
        \includegraphics[width=1.0\linewidth]{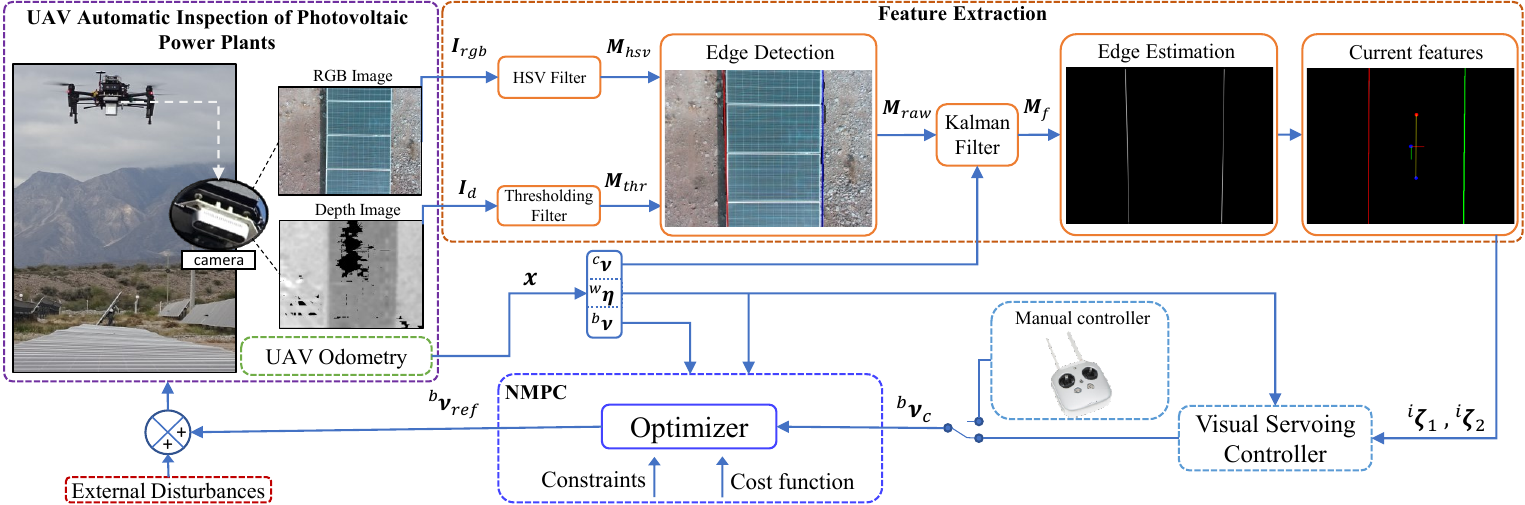}
    \caption{Pipeline of the system implemented for overflight and inspection of the PV arrays with the visual servoing controller combined with dynamic compensation and constraints based on NMPC using UAVs. The purple area shows the UAV system in conjunction with the test environment. Visual servo controllers and NMPC are explained in Algorithms \ref{alg:knematics} and \ref{alg:nmpc}, respectively. The orange area shows the Feature Extraction from the RGB-D images and is explained in Algorithm \ref{alg:kalman_filter}).}
    \label{fig:pipeline_of_VS-NS-MPC}
\end{figure*}

Unmanned aerial vehicle (UAV) technology has experienced accelerated development. It has made a significant contribution to monitoring applications such as oil and pipeline inspection \cite{Aljalaud2023AutonomousBehavior}, power transmission lines \cite{Xu2022PowerSystem}, and precision agriculture \cite{MUHAMMADNAVEEDTAHIR202355}. This technology has recently been applied to PV inspection, saving valuable time and human resources, as well as ensuring rapid data collection over large areas \cite{ Aghaei2016ImageTechnologies,Li2017VisibleSystems, KirstenVidaldeOliveira2020AerialPlants,Zefri2022DevelopingData,SegoviaRamirez2022UnmannedPanels}.
Currently, most of the UAV applications in PV inspection are based on photogrammetry, where UAVs are equipped with a Global Positioning System (GPS) receiver and an Inertial Measurement Unit (IMU) \cite{Zefri2021In}. These elements ensure that the system flies at high altitude from the ground, guaranteeing proper horizontal and vertical overlap of the images for orthomosaic reconstruction \cite{Zefri2022AppliedDevelopment, Hernandez-Lopez2023SunMap:Photogrammetry}. 

Nevertheless, the photogrammetry approach has two limitations. First, GPS accuracy can lead to incorrect placement of the UAV, resulting in an increased amount of useless data during flights. Second, flights at high altitude from the ground can generate issues related to image resolution, negatively affecting the surface defects detection process. The results presented in \cite{Aghaei2016ImageTechnologies} have shown an intrinsic relationship between the ground sampling distance and the detection of surface defects, such as soil, snail trails, and white spots. Thus, resolution is a key factor in achieving accurate detection of surface failures. Therefore, we find it interesting to focus our research here, making it an important contribution to the field of PV inspections.

In this paper, we propose an approach to ensure proper image acquisition of PV modules based on autonomous UAVs, which is capable of capturing details on the surface of the PV array without geo-referenced measurements or path planning methods. The objective of this work is to develop a visual servoing control system combined with a Nonlinear Model Predictive Control (NMPC) capable of accurately tracking the center of the underlying PV array at low-altitude. This system ensures the acquisition of images for later use in the inspection process by experts or AI-based architectures. Fig. \ref{fig:pipeline_of_VS-NS-MPC} shows the pipeline of our approach.

In summary, our contributions are the following.
\begin{itemize}
    \item  A visual servoing control system for aerial vehicles combined with dynamic compensation and constraints based on NMPC, which ensures accurate tracking of the middle of the PV arrays considering different frontal velocities and height constraints without geo-referenced points or path planning methods.

   \item Lightweight and low run-time PV panel feature extraction method based on RGB-D images and a Kalman filter to mitigate the impact of noise and uncertainties.
    
   \item End-to-end validation of the complete system using a commercial aerial vehicle (DJI Matrice 100), both in a simulated environment and in real-world conditions.  
  
\end{itemize}

\section{Methodology}
\label{sec:Methodology}

\subsection{Notation}
\label{subsec:Notation}
Throughout the article, we establish the following definitions: an inertia frame denoted as $\mathcal{W} = \lbrace O_\mathcal{W}, W_x, W_y, W_z \rbrace$, and the body frame denoted by $\mathcal{B}= \lbrace O_\mathcal{B}, B_x, B_y, B_z \rbrace$ located at  the UAV’s center of mass (CoM). Furthermore, the camera frame can be defined as \mbox{$\mathcal{C} = \lbrace O_\mathcal{C}, C_x, C_y, C_z \rbrace$}, where $O_\mathcal{C}$ aligns with the optical center of the camera and with the $2D$ digital image plane defined as $\mathcal{I} = \lbrace O_\mathcal{I}, I_u, I_v \rbrace$ (see Fig. \ref{fig:drone_axes_and_features}). Moreover, this work uses the symbol $^{\mathcal{B}}\mathbf{R}_\mathcal{C}$ to denote a rotational matrix from the frame $\mathcal{C}$ to $\mathcal{B}$. Vectors are represented by bold quantities as follows: $^{w}\mathbf{p} \in \mathbb{R}^{n}$, using the frame ${\mathcal{W}}$ as a prefix; matrices are indicated in bold capital letters as $\mathbf{P} \in \mathbb{R}^{n \times m}$. Finally, the $\mathbf{Q}$-weighted norm can be formulated as $\| \mathbf{p} \|_{\mathbf{Q}} = \sqrt{\mathbf{p}^T \mathbf{Q} \mathbf{p}}$

\subsection{System modeling}
\label{subsec:System modeling}
This section presents the formulation of motion equations considering the UAV and the camera. These equations will serve as the foundation for the visual servoing NMPC formulation applied to the aerial vehicle. Additionally, image feature extraction is carefully formulated to ensure accurate extraction, leveraging the combination of RGB-D information with a Kalman filter.
\vspace{2mm}
\subsubsection{Quadrotor Differential Kinematics}
\begin{figure}[ht]
    \centering
        \includegraphics[width=1.0\linewidth]{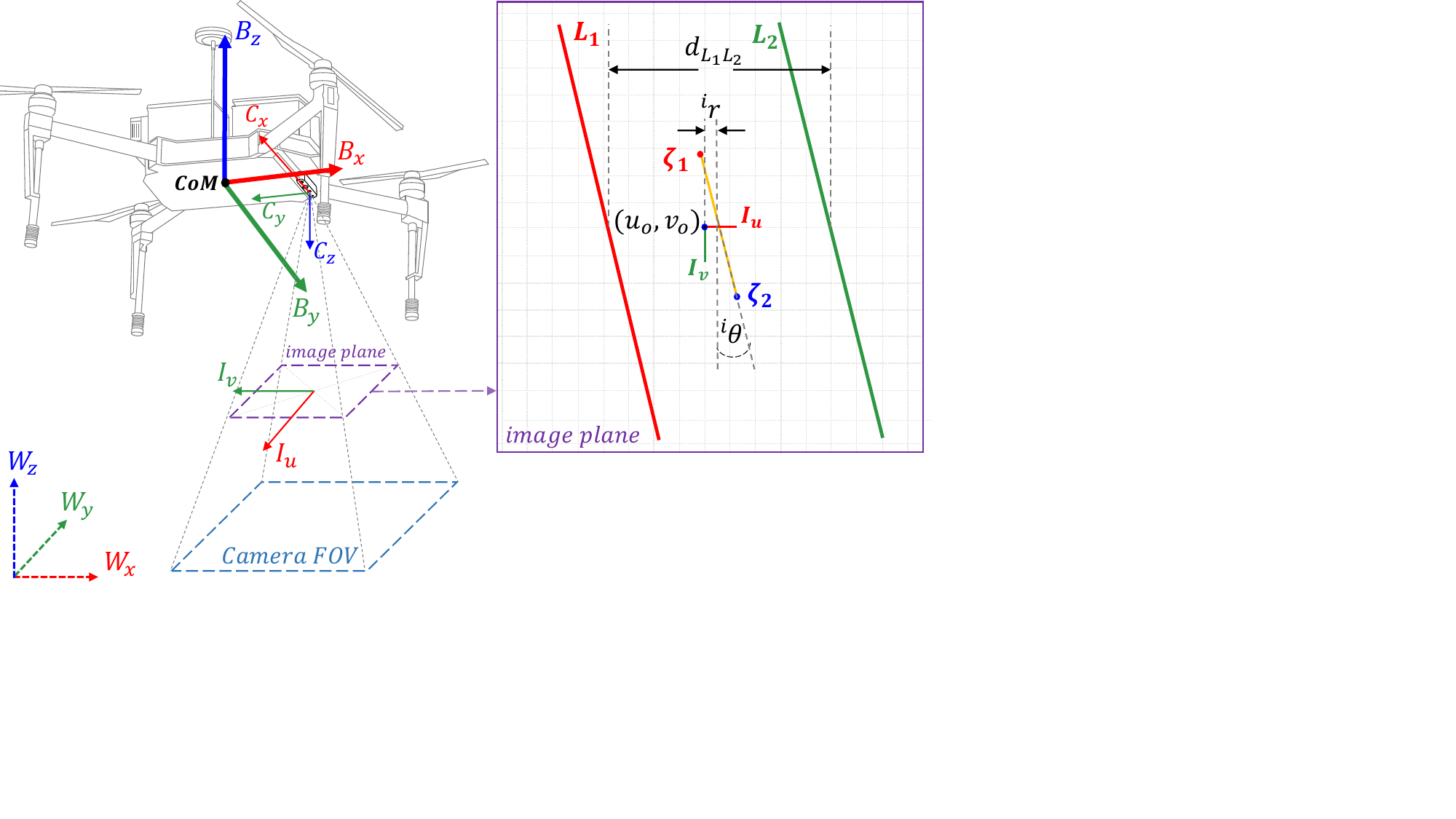}
    \caption{Reference coordinate frames of the implemented system.}
    \label{fig:drone_axes_and_features}
\end{figure}

This work formulates the equations of motion, considering that the aerial system (DJI Matrice 100) includes an autopilot that ensures tracking of the reference velocities with respect to the frame $\mathcal{B}$. This consideration takes into account that the roll angle $\phi \approx 0$ and pitch angle $\theta \approx 0$ are approximately zero.
Therefore, the location of the camera $^{w}{\bm{\eta}}$ can be formulated as follows:
\begin{equation}
    \begin{bmatrix}
 ^{w}x \\  ^{w}y \\  ^{w}z \\  ^{w}{\psi}  \end{bmatrix} =  \begin{bmatrix} ^{w}{x_b} + ~^b x_c \cos({\psi}) -~^b y_c \sin({\psi}) \\ ^{w}{y_b}  + ~^b x_c \sin({\psi})
               +~^b y_c \cos({\psi}) \\ ^{w}{z_b} + ~^b z_c\\ \psi \end{bmatrix}
    \label{eq:kinematics:1}
\end{equation}
where $(^{w}{x_b}, ^{w}{y_b}, ^{w}{z_b})$ is the location of frame $\mathcal{B}$  respect to the frame $\mathcal{W}$, $(~^b x_c, ~^b y_c, ~^b z_c)$ is the camera's position respect to the frame $\mathcal{B}$; finally, the rotational angle with respect to axis $W_z$ can be expressed as $\psi$ (for more details, see the reference coordinate frames in Fig. \ref{fig:drone_axes_and_features}).

Taking it into account and based on \cite{RECALDE2022353}, the differential kinematics can be formulated as:
\begin{equation} \label{eq:kinematics:2}
^{w}\dot{\bm{\eta}}=\mathbf{J}_b(\psi) ^{b}\bm{\nu}
\end{equation}

where $\mathbf{J}_b \in \mathbb{R}^{4 \times 4}$ is the Jacobian matrix that allows the liner mapping of the UAV's velocities \mbox{$^{b}\bm{\nu}= \begin{bmatrix} ^{b}{\upsilon}_{x} & ^{b}{\upsilon}_{y} & ^{b}{\upsilon}_{z} & ^{b}{\omega}_{z} \end{bmatrix}^{T} \in \mathbb{R}^4$} considering that the angular velocities $^{b}{\omega}_{x}$ and $^{b}{\omega}_{y}$ are approximately zero due to the autopilot of the system.
\vspace{5px}

\subsubsection{Quadrotor Dynamics}

Taking the presented autopilot into account and based on \cite{drones7020144}, the dynamic model can be formulated as follows:

\begin{equation} \label{eq:dynamics:1}
 ^{b}\bm{\nu}_{ref} = \mathbf{{M}}(^{b}\bm{\nu},\bm\pi) ^{b}\dot{\bm{\nu}} + \mathbf{{C}}(^{b}\bm{\nu},\bm\pi)^{b}\bm{\nu}
\end{equation} 
where $\mathbf{{M}} \in \mathbb{R}^{4 \times 4}$ and $\mathbf{{C}} \in \mathbb{R}^{4 \times 4}$ represent the matrices for inertial and coriolis forces and $^{b}\bm{\nu}_{ref}$ is the vector of desired velocities of the UAV. Additionally, $\bm\pi \in \mathbb{R}^{18}$ is a vector used to characterize the dynamics of the system.
\vspace{2mm}
\subsubsection{Image Features}
The pinhole camera model is considered in this work \cite{Keipour2022VisualVehicle}, where the  feature point $^{c}\bm{\zeta} = \begin{bmatrix}
^{c}x & ^{c}y &  ^{c}z\end{bmatrix}^{T}  \in \mathbb{R}^3$ expressed in $\mathcal{C}$ can be projected on the digital image plane as follows:

\begin{equation}
    \begin{bmatrix}
^{i}u \\ ^{i}v  \end{bmatrix} = \frac{\lambda}{~^{c}z} \begin{bmatrix} ~^{c}x \\ ~^{c}y \end{bmatrix}
    \label{eq:camera:1}
\end{equation}
where $^{i}\bm{\zeta} = \begin{bmatrix}
^{i}u & ^{i}v \end{bmatrix}^{T}  \in \mathbb{R}^2$ is the feature vector projected into the digital image plane  $\mathcal{I}$ and $\lambda$ is the camera focal length.

Our work considers a fixed feature point $^{i}\dot{\bm{\zeta}} = \mathbf{0}$; therefore, the temporal variation in the image plane can be expressed in relation to the velocities of the UAV, resulting in the following expression:
\begin{equation}
    ^{i}\dot{\bm{\zeta}}= \mathbf{J}_f(^{i}\bm{\zeta},~^{c}z) \mathbf{T}^{-1}~^{b}\bm{\nu}
 \label{eq:camera:3}
 , \:\: \mathbf{T} = \begin{bmatrix} {^{\mathcal{B}}\mathbf{R}_{\mathcal{C}}} & [^b\mathbf{t}_c]_\times {^{\mathcal{B}}\mathbf{R}_{\mathcal{C}}} \\ \mathbf{0}_{3 \times 3} & {^{\mathcal{B}}\mathbf{R}_{\mathcal{C}}} \end{bmatrix}
 \end{equation}

where $^{b}\bm{\nu} = \begin{bmatrix}
{^{b}\bm{\upsilon}} & {^{b}\bm{\omega}} \end{bmatrix}^{T}  \in \mathbb{R}^6$ is the stacked vector of the UAV's velocities without the consideration of the autopilot, $[^b\mathbf{t}_c]_\times$ is a  skew symmetric matrix associated to \hbox{$^b\mathbf{t}_c = \begin{bmatrix} ~^b x_c & ~^b y_c  & ~^b z_c \end{bmatrix}^{T} \in \mathbb{R}^{3}$}. Finally, considering two features vectors $^{i}\bm{\zeta}$, the Jacobian image matrix is $\mathbf{J}_{f} \in \mathbb{R}^{4 \times 6}$.
\vspace{2mm}
\subsubsection{Line Features}
\label{subsec:line_feature}
This section presents the formulation of the line features expressed by two points that characterize the center of the PV array. These points are projected on the digital image plane as  $^{i}\bm{\zeta}_1 = \begin{bmatrix}
^{i}u_1 & ^{i}v_1 \end{bmatrix}^{T}$ and $^{i}\bm{\zeta}_2 = \begin{bmatrix}
^{i}u_2 & ^{i}v_2 \end{bmatrix}^{T}$ presented in Fig. \ref{fig:drone_axes_and_features} (more information related to the extraction of the two points can be found in \ref{subsec:Image features extraction}). This work further formulates these two points in polar coordinates as:

\begin{equation}
    \begin{bmatrix}
^{i}r \\ ^{i}\theta  \end{bmatrix} =  \begin{bmatrix} \sin{(^{i}\theta)} (^{i}u_{1} - (\frac{^{i}u_2 - ^{i}u_1}{^{i}v_2-^{i}v_1}) ~^{i}v_{1}) \\ \arctan{(-\frac{^{i}v_2 - ^{i}v_1}{^{i}u_2-^{i}u_1} )} \end{bmatrix}
    \label{eq:camera:4}
\end{equation}
where $^{i}\bm{\xi} = \begin{bmatrix}
^{i}r & ^{i}\theta \end{bmatrix}^{T}  \in \mathbb{R}^2$ is the line features vector, which characterize the middle line of the lateral edges of the PV panels respect to the frame $\mathcal{I}$.

The time derivative of \eqref{eq:camera:4} can be formulated as a function of the camera velocities by combining \eqref{eq:camera:3}, defining the following formulation:
\begin{equation}
    ^{i}\dot{\bm{\xi}}= \mathbf{J}(^{i}\bm{\xi},^{i}\bm{\zeta},{}^{c}z)^{b}\bm{\nu}
 \label{eq:camera:5}
  \end{equation}
 $$
 \mathbf{J}= \mathbf{J}_l(^{i}\bm{\xi},^{i}\bm{\zeta})\mathbf{J}_f(^{i}\bm{\zeta},{}^{c}z)\mathbf{T}^{-1}
 $$
where, $\mathbf{J}_l \in \mathbb{R}^{2 \times 4}$ is the Jacobian matrix associated with the line features, and $\mathbf{J} \in \mathbb{R}^{2 \times 6}$ is a compact Jacobian, which includes $\mathbf{J}_l$ and $\mathbf{J}_{f}$ for simple notation.
The formulation \eqref{eq:camera:5} is suitable for developing the visual servoing controller as presented in \ref{subsec:Visual servoing system}, without considering the dynamics of the UAV.

\subsection{Visual Servoing Control System Combined with NMPC}
\label{subsec:Visual servoing system}
\subsubsection{Visual Servoing Controller}
The visual servoing controller proposed in this work aims to guide the UAV in a way that the feature points of the line $^{i}\bm{\xi}$ move towards the desired values $^{i}\bm{\xi}_{d} = \begin{bmatrix} ^{i}r_{d} & ^{i}\theta_{d} \end{bmatrix}^{T}$, ensuring the movement of the UAV over the middle of the PV arrays. Taking this into account, the control error vector can be formulated as $^{i}\Tilde{\bm{\xi}} = ~^{i}\bm{\xi}_{d} - \alpha(^{i}\bm{\xi})$; however, due to $|^{i}r| \gg |^{i}\theta|$, the operator $\alpha(^{i}r , ^{i}\theta) = \begin{bmatrix} ^{i}r/^{i}r_{max} &  ^{i}\theta\end{bmatrix}^T$ ensures that the values are of the same magnitude order, considering that $^{i}r_{\text{max}}$ represents half of the image width.
Due to the dimensions of the Jacobian matrix $\mathbf{J} \in \mathbb{R}^{2 \times 6}$, there is no unique solution, and it is possible to use the null space projector operator. Furthermore, when considering the nearly perfect tracking velocity $^{b}\bm{\nu} \equiv ~^{b}\bm{\nu}_{c}$, the control law can be formulated as:

\begin{equation} 
^{b}\bm{\nu}_{c} = \mathbf{J}^\dag (~^{i}\dot{\bm{\xi}_d} +\mathbf{K}^{i}\Tilde{\bm{\xi}})+ (\mathbf{I}_{6 \times 6}-\mathbf{J}^\dag\mathbf{J})^{b}\bm{\nu}_d\label{eq:camera:6}
\end{equation}
note that $\mathbf{J}^\dag = \mathbf{W}^{-1}\mathbf{J}^T(\mathbf{J}\mathbf{W}^{-1}\mathbf{J}^{T})^{-1}$ is the Moore-Penrose pseudoinverse, $(\mathbf{W}, \mathbf{K})$ are positive definite matrices and $(\mathbf{I}_{6 \times 6}-\mathbf{J}^\dag\mathbf{J})$ is the orthogonal projection of  $^{b}\bm{\nu}_d$.

The vector $^{b}\bm{\nu}_d$  contains arbitrary values for the associated velocities of the UAV in the frame ${\mathcal{B}}$ with \mbox{$^{b}\bm{\nu}_d= \begin{bmatrix} ^{b}{\upsilon}_{xd} & ^{b}{\upsilon}_{yd} & ^{b}{\upsilon}_{zd} & ^{b}{\omega}_{xd} & ^{b}{\omega}_{yd} & ^{b}{\omega}_{zd} \end{bmatrix}^{T}$}. The appropriate selection of $\mathbf{W}$ can guarantee angular velocities $^{b}{\omega}_{xd}\approx 0$ and $^{b}{\omega}_{yd}\approx 0$, making the control proposal compatible with the aerial system (DJI Matrice 100).
The arbitrarily long vector can be formulated in order to guarantee any frontal velocity and height over the PV array, which can be written as follows:

\begin{equation}
    \begin{aligned}
    ^{b}{\upsilon}_{xd} &=  \frac{{~^{b}\upsilon}^{max}_{x}}{1 + k_1 ||^{i}\Tilde{\bm{\xi}}||} \\ ~^{b}{\upsilon}_{zd} &= ~\dot{\eta}_{zd} + k_2 \tanh{(^w\eta_{zd} - ~^w\eta_z)}
    \end{aligned}
    \label{eq:camera:7}
\end{equation}
where ${~^{b}\upsilon}^{max}_{x}$ is the maximum desired frontal velocity, $(k_1, k_2)$ are positive scalar values and $~^w\eta_{zd}$ represents the desired height of the UAV over the PV solar panels. In our application, $~\dot{\eta}_{zd}=0$ and $^{i}\dot{\bm{\xi}_d}=0$, due to the fact that the flights over the PV arrays are maintained at a constant height and the desired line features are constant over time.\\
Algorithm \ref{alg:knematics} explains the procedure for develop the visual servoing controller.

\begin{algorithm}[ht]
\footnotesize	
\label{alg:knematics}
\caption{Visual-Servoing}
\KwData{$^{i}\bm{\xi}_{d}$, $^{i}\dot{\bm{\xi}_d}$, $^{i}\bm{\zeta}_1$, $^{i}\bm{\zeta}_2$, $^{w}{\bm{\eta}}$} 
\KwResult{$^{b}\bm{\nu}_{c}$ Control value}
\SetAlgoLined
$^{i}\bm{\xi}$ $\gets$ $^{i}\bm{\zeta}_1$,  $^{i}\bm{\zeta}_2$; // \textit{Image Features} \\
$^{i}\Tilde{\bm{\xi}} = ~^{i}\bm{\xi}_{d} - \alpha(^{i}\bm{\xi})$;  // \textit{Control Error} \\  
$^{b}{\upsilon}_{xd} = {~^{b}\upsilon}^{max}_{x} / (1 + k_1 ||^{i}\Tilde{\bm{\xi}}||)$; // \textit{Desired frontal Velocity} \\  
$^{b}{\upsilon}_{zd} = ~\dot{\eta}_{zd} + k_2 \tanh{(^w\eta_{zd} - ~^w\eta_z)}$; // \hbox{\textit{Desired upper Velocity}} \\  
$^{b}\bm{\nu}_d \gets ^{b}{\upsilon}_{xd}, ~^{b}{\upsilon}_{zd}$; // \hbox{\textit{Arbitrary velocity vector}} \\ 
 $\mathbf{J} \gets ~^{i}\bm{\xi}, ~^{i}\bm{\zeta}_1, ~^{i}\bm{\zeta}_2, \mathbf{T}$; // \hbox{\textit{Compact Jacobian}} \\ 
$^{b}\bm{\nu}_{c} = \mathbf{J}^\dag (~^{i}\dot{\bm{\xi}}_d +\mathbf{K}^{i}\Tilde{\bm{\xi}})+ (\mathbf{I}_{6 \times 6}-\mathbf{J}^\dag\mathbf{J})^{b}\bm{\nu}_d$; // \textit{Control Law} \\ 

\end{algorithm}

The proposal \eqref{eq:camera:6} does not consider the dynamics of the system and lacks information about constraints related to maximum velocities or the workspace within which the system can ensure adequate feature extraction. Consequently, there is a need to formulate an NMPC that incorporates these constraints.
\vspace{2mm}
\subsubsection{Dynamic Compensation and Constraints with NMPC}

This section introduces a NMPC approach to ensure a nearly perfect tracking velocity of the control actions generated by the visual servoing controller $^{b}\bm{\nu} \equiv ~^{b}\bm{\nu}_{c}$. This NMPC is designed to compute the control policy while incorporating the dynamics of the system and the restricted operational space as constraints in a finite-time optimal control problem (OCP). Thus, this work utilizes a discrete version of equations \eqref{eq:kinematics:2} and \eqref{eq:dynamics:1}. These equations were formulated using the fourth-order Runge-Kutta integrator, considering $dt$ as the sampling time, defined as follows:
\begin{equation}\label{eq:mpc:1}
   {\mathbf{x}}_{k+1} = \mathbf{f}_{RK4}(\mathbf{f}(\mathbf{x}_k,\mathbf{u}_k), dt)
\end{equation} 
$$
\mathbf{f}(\mathbf{x},\mathbf{u})~= \begin{bmatrix}
\mathbf{J}_b(\psi) ~^{b}\bm{\nu} \\
-(\mathbf{{M}}^{-1}\mathbf{{C}})~^{b}\bm{\nu} + \mathbf{{M}}^{-1} ~^{b}\bm{\nu}_{ref}
\end{bmatrix}
$$
where $\mathbf{x} = \begin{bmatrix} ^{w}{\bm{\eta}} & ~^{b}\bm{\nu} \end{bmatrix}^T  \in \mathbb{R}^{8} $ and $\mathbf{u} = \begin{bmatrix}~^{b}\bm{\nu}_{ref}  \end{bmatrix}^T \in \mathbb{R}^{4}$ are the states and control actions of the system.

The intermediate cost function related to velocity tracking performance $l_v(\cdot): \mathbb{R}^{4}  \rightarrow \mathbb{R}_{\geq 0}$ can be formulated as follows:

\begin{equation} \label{eq:mpc:2}
 \begin{aligned}
 l_v(\mathbf{x}_k, ~^{b}\bm{\nu}_{c_k}) = \Vert  ^{b}\bm{\nu}_{c_k}-\mathbf{C}_v\mathbf{x}_k\Vert_{\mathbf{Q}}
     \end{aligned}
\end{equation}

where $^{b}\bm{\nu}_{c}$ is the control action generated by the visual servoing controller \eqref{eq:camera:6}, $\mathbf{C}_v$ is a matrix designed to obtain the velocities of the aerial system.
Finally, $\mathbf{Q}$ is a positive definite matrix. Moreover, to ensure smooth control actions, the cost function associated with the control policy $l_u(\cdot): \mathbb{R}^{4}  \rightarrow \mathbb{R}_{\geq 0}$  can be structured: $l_{u}(\mathbf{u}_{k}) = \Vert  \mathbf{u}_{k}\Vert_{\mathbf{R}}$, where ${\mathbf{R}}$ is a positive definite matrix.

In addition to the equations of motion for the aerial system, constraints related to control actions and workspace can be formulated as $\mathbf{x}_{min} \leq \mathbf{x}_k \leq \mathbf{x}_{max}, \: \mathbf{u}_{min} \leq \mathbf{u}_k \leq \mathbf{u}_{max}$. $(\mathbf{x}_{min}, \mathbf{x}_{max}, \mathbf{u}_{min},\mathbf{u}_{max})$ represents the minimum and maximum states and control actions of the aerial system. These constraints ensure proper velocities and height compliance, providing the acquisition of detailed images and adequate feature extraction during low-altitude flights over the PV modules. Therefore, the cost functions presented previously, it is possible to generate the NMPC as a finite-time horizon optimization problem as follows:

\begin{equation}\label{eq:mpc:5}
\begin{split}
     \argminA_{\small\begin{matrix} \mathbf{x}_1 \hdots \mathbf{x}_N \\
         \mathbf{u}_1 \hdots \mathbf{u}_{N-1}
     \end{matrix}}  \quad  l_v(\mathbf{x}_N, ~^{b}\bm{\nu}_{c_N})
 + \sum_{k=1}^{N-1} l_v(\mathbf{x}_k, ^{b}\bm{\nu}_{c_k}) +  l_{u}(\mathbf{u}_{k}) \\  
    \textrm{subject to: }  {\mathbf{x}}_{k+1} = \mathbf{f}_{RK4}(\mathbf{x}_{k},\mathbf{u}_{k}), k = 1,., N-1 ~~~~~\\
    \mathbf{u}_{min} \leq \mathbf{u}_k \leq \mathbf{u}_{max},  k = 1,., N-1 ~~~~~~~~\\
    \mathbf{x}_{min} \leq \mathbf{x}_{k} \leq \mathbf{x}_{max},  k = 1,., N ~~~~~~~
\end{split}
\end{equation}

NMPC is solved using Sequential Quadratic Programming (SQP) in a real-time iteration (RTI) scheme using ACADOS \cite{Verschueren2022AcadosaControl} and CasADi \cite{Andersson2019CasADiaControl}. The Algorithm \ref{alg:nmpc} comprehensively explains the Visual Servoing controller combined with the NMPC formulation.

\begin{algorithm}
\footnotesize	
\label{alg:nmpc}
\caption{Visual-Servoing + NMPC}
\KwData{$^{i}\bm{\xi}_{d}$, $^{w}{\bm{\eta}}$, $^{b}\bm{\nu}$ Desired line features and Internal states of the system} 
\SetAlgoLined
\LinesNumbered
\While{\text{Inspection is not complete}}{
    $\mathbf{x} \gets ~^{w}{\bm{\eta}}, ~^{b}\bm{\nu}$; // \textit{Get internal states of the UAV}\\    
    $^{i}\bm{\zeta}_1$, $^{i}\bm{\zeta}_2$ $\gets$ \text{Image features Extraction};\\   
    $^{b}\bm{\nu}_{c}$ $\gets$ \text{Visual-Servoing}\:($^{i}\bm{\xi}_{d}$, $^{i}\bm{\zeta}_1$, $^{i}\bm{\zeta}_2$, $^{w}{\bm{\eta}}$)\;     
     $^{b}\bm{\nu}_{ref}$ $\gets$ \text{NMPC}\:($\mathbf{x}$, $^{b}\bm{\nu}$, $^{b}\bm{\nu}_{c}$); // \textit{ACADOS Optimizer}\\
     \text{APPLY}($^{b}\bm{\nu}_{ref}$) \text{to the UAV}
}
\end{algorithm}

\subsection{Image features extraction}
\label{subsec:Image features extraction}
We propose a feature extraction of the PV array by approximating to two straight lines the lateral edges of the photovoltaic panels. By identifying the lateral edges of the PV array, we extract the deviation with respect to the center of the image $(^{i}r)$ and the angle of inclination of the parallel lines generated by these edges $(^{i}\theta)$. We prolong the line with the average of the slopes of $L_1$ and $L_2$ using the midpoint (see image plane in Fig. \ref{fig:drone_axes_and_features}). Once the new slope is calculated, we determine the $u,v$ components of the points $^{i}\zeta_1$ and $^{i}\zeta_2$ as follows:

\begin{equation}
   \begin{matrix}
^{i}\zeta_{u_{1,2}} = (u_o \pm p_f cos(^{i}\theta))
\\
^{i}\zeta_{v_{1,2}} = (v_o \pm p_f sin(^{i}\theta))
\end{matrix}
    \label{eq:pixels_on_line}
\end{equation}

where $p_f$ is a factor that depends on the horizontal distance $d_{L_1L_2}$, and the desired distance at which the camera wants to capture the lines $L_1$ and $L_2$ of the PV array.
\vspace{2mm}
\subsubsection{Line Detection} 

To detect the PV array's vertical lines, RGB images undergo HSV filtering, highlighting white edges, while depth images are thresholded based on camera-to-PV and camera-to-ground distances. Thus, the RGB and D images ($I_{rgb}, I_d$) are binarized, noise-filtered with morphological operations, and merged. Then, we generate a $M_{raw}$ mask that approximates straight lines using OpenCV's \textit{fitline} algorithm.

We employ a Kalman filter to estimate the states $\mathbf{s}_k$ derived from the \textit{fitline} algorithm, influenced by camera velocities $^{c}\bm{\nu}$. The matrices $\mathbf{A}$ and $\mathbf{B}$ are determined by the dynamic mode decomposition (DMD) methodology \cite{Bai2020DynamicIdentification}. We perform a manual flight over the PV array, varying UAV velocities, altitudes, and attitude. This yields the data set for states $\mathbf{s}_k$ and system inputs $^{c}\bm{\nu}$ used in the estimation of $\mathbf{A}$ and $\mathbf{B}$.

\begin{algorithm}
\footnotesize	
\label{alg:kalman_filter}
\caption{Image feature extraction}
\KwData{ $^c\bm{\nu}$, $I_{rgb}$ and $I_{d}$ image }
\KwResult {$^{i}\bm{\zeta}_1$, $^{i}\bm{\zeta}_2$ Feature extraction}
\SetAlgoLined
\LinesNumbered
Load the RGB and Depth image\;
\While{$(I_{rgb} \And I_{d}) \neq \varnothing$}{
    $M_{hsv} \leftarrow I_{rgb} \rightarrow HSV_{filter}$\; 
    $M_{thr} \leftarrow I_{d} \rightarrow Thresholding_{filter}$\; 
    $M_{aux} \leftarrow M_{hsv} \cup M_{thr}$\;
    $M_{raw} \leftarrow M_{aux} \rightarrow  Morphological_{filter}$\;
    \textbf{Vertical contours extraction Loop}
    \hspace{1em}$(contour_{left}\; , \;contour_{rigth}) \leftarrow M_{raw}$\;
    \textbf{end Loop}
    \hspace{1em} $x_k \leftarrow \textit{fitline}(contour_{left}\; , \;contour_{rigth})$\;  \textit{// Kalman Filter} \\
    \hspace{1em} $\mathbf{s}_{k+1} = \mathbf{A} \hat{\mathbf{s}}_{k-1} + \mathbf{B}\; ^c\bm{\nu} $\;
    \hspace{1em} $\hat{\mathbf{P}}_k = \mathbf{A}\mathbf{P}_k \mathbf{A}^{T}+ \mathbf{Q}_k $\;
    \hspace{1em} $\mathbf{K}_{k} = \mathbf{P}_{k+1}\mathbf{H}^T (\mathbf{H}\mathbf{P}_{k+1}\mathbf{H}^T + \mathbf{R})^{-1}$\;    
    \hspace{1em} $\hat{\mathbf{s}}_k = \mathbf{s}_{k+1} + \mathbf{K}_{k} (\mathbf{s}_{k}-\mathbf{H} \mathbf{s}_{k+1})$; // \textit{Estimation}\\
    \hspace{1em} $\mathbf{P}_k = (\mathbf{I}_{4\times4} - \mathbf{K}_{k} \mathbf{H})\hat{\mathbf{P}}_k$\;
    \hspace{1em} $ M_f \leftarrow \hat{\mathbf{s}}_k$; // \textit{New mask filtered} \\
    $^{i}\bm{\zeta}_1$, $^{i}\bm{\zeta}_2$ $\gets$  $M_f$; // \textit{Feature extraction of Mask filtered}
}
\end{algorithm}

Algorithm \ref{alg:kalman_filter} describes the procedure developed for feature extraction, taking RGB-D images $I_{rgb}$, $I_{d}$ and camera velocities $^c\bm{\nu}$ as inputs. The output is the $M_{f}$ mask representing the current features based on the estimated parameters of the lines $\hat{\mathbf{s}}_k$, which correspond to the edges of the PV array.

\section{Experiments and results}
\label{sec:Experiments and results}

This section showcases our method's performance in simulated and real experiments. We outline the experimental setups for the simulation environment and the real platform, including gain weight settings and the controller used. The controller gains were tuned through experimentation, initially in the simulator and later fine-tuned in real experiments.

\subsection{Experimental setup}
\label{sec:Experimental setup}
\subsubsection{Simulation Experiments}
We utilized Webots \cite{michel_2004} to simulate PV arrays and DJI Assistant to ensure a similar behavior of the UAV Matrice 100 (see \mbox{Fig. \ref{fig:dron_simulador_real}}). Webots includes a simulated environment with a RGB-D camera mounted on the UAV. Both simulators communicate via topics using ROS Noetic, allowing integration of the flight dynamics in DJI Assistant with the Webots simulation environment. Furthermore, we considered noise in sensors, with the following Gaussian distribution $^{w}{\bm{\eta}}_n \sim \mathcal{N} (-0.005,0.005)$ and $~^{b}\bm{\nu}_n \sim \mathcal{N} (-0.001,0.001)$. The simulators were implemented on a computer with an {AMD\textregistered} Ryzen 7 3700x 8-core processor, 16 GB of RAM and a NVIDIA GTX 1060 video card with 6 GB of memory, using the Ubuntu 20.04 operating system.

\subsubsection{Real Experiments} 
\begin{figure}[ht]
    \centering
        \includegraphics[width=1.0\linewidth]{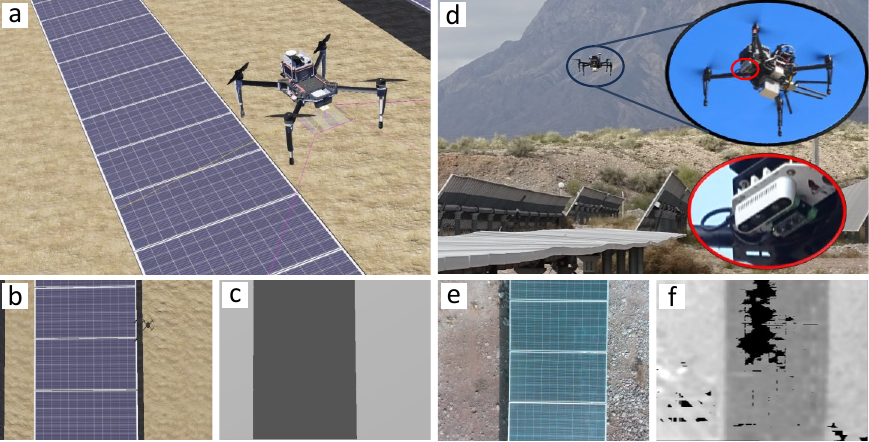}
    \caption{UAV flying over the PV arrays. Figures \textbf{a} and \textbf{d} represent the simulated and real experimental environment, respectively. The bottom images \textbf{b}, \textbf{c} and \textbf{d}, \textbf{f} are the color and depth images of the PV array obtained by simulation as well as by the UAV on-board camera.}
    \label{fig:dron_simulador_real}
\end{figure}

We used a DJI Matrice 100 drone in combination with a Jetson Orin-NX onboard computer, equipped with \hbox{16 GB RAM} and running the Ubuntu 20.04 Focal operating system with Jetpack 5.1.1. As RGB-D sensor, we used an {Intel\textregistered} {RealSense\texttrademark} D435i camera. The real experiments were carried out at the Anchipurac Photovoltaic Solar Park, located in the city of San Juan, Argentina (Fig. \ref{fig:dron_simulador_real}). In this environment, we used a PV array of 80 solar panels with individual dimensions of 2.0 x 1.0 $[m]$. 

In the simulation and real experiments, we used images with a resolution of 640x480 at 30 \textit{fps} and the PV arrays were placed horizontally so that the aerial images fully displayed each solar panel. Furthermore, \hbox{Table \ref{table:2}} shows the values of all parameters explained in Section \ref{sec:Methodology}. 

\subsection{Simulation experiment results}\label{subsec:Simulation experiment results}

The first experiment aimed to determine the optimal matrix $\mathbf{W}$ values for the visual servoing controller \eqref{eq:camera:6}, crucial for the proposed controller. Simulating a scenario where the UAV camera maintains a direct line of sight to the photovoltaic array, we ensured ideal feature extraction with specified values in the null space projector: \hbox{${~^{b}\upsilon}^{max}_{x} = 1~[m/s]$} and $^w\eta_{zd} = 3~[m]$ over the PV arrays.

\begin{table}[h]
    \caption{Visual Servoing and NMPC Parameters} 
    \renewcommand{\arraystretch}{1.2}
    \begin{adjustbox}{width=\columnwidth,center}
    \begin{tabular}{c c c c} 
        \hline\hline
         {\bfseries Parameters} & {\bfseries Values} & {\bfseries Parameters} & {\bfseries Values} \\ [1ex] 
        \hline 
        $\mathbf{K}$ & $diag(150~0.5)$& $\mathbf{W}$ & $diag(1~1~1~50~50~1)$ \\[0.5ex]
        $k_1 $ & $10$& $k_2$ & $10$ \\[0.5ex]
        $\mathbf{Q}$ & $diag(1~1~1~1)$& $\mathbf{R}$ & $diag(1.4~1.4~1.4~1.4)$ \\[0.5ex]
        $\mathbf{x}_{min}$ & $-(\infty~\infty~0~\pi~2~2~2~1.5)$& $\mathbf{x}_{max}$ & $(\infty~\infty~4.5~\pi~2~2~2~1.5)$\\[0.5ex]
        $\mathbf{u}_{min}$ & $-(2~2~2~1.5)$& $\mathbf{u}_{max}$ & $(2~2~2~1.5)$\\[0.5ex]
        $^{i}r_{d}$ & $0$& $^{i}\theta_{d}$ & $0~[rad]$ \\[0.5ex]
        $N$ & $10$& $dt$ & $0.05~[s]$\\[0.5ex]
        \hline 
    \end{tabular}
    \end{adjustbox}
    \label{table:2} 
\end{table}

\begin{figure}[ht]
    \centering
    \subfloat[Inappropriate selection of matrix $\mathbf{W}$ \label{fig:simulation_experiments_findW_a}]{%
        \includegraphics[width=0.5275\linewidth]{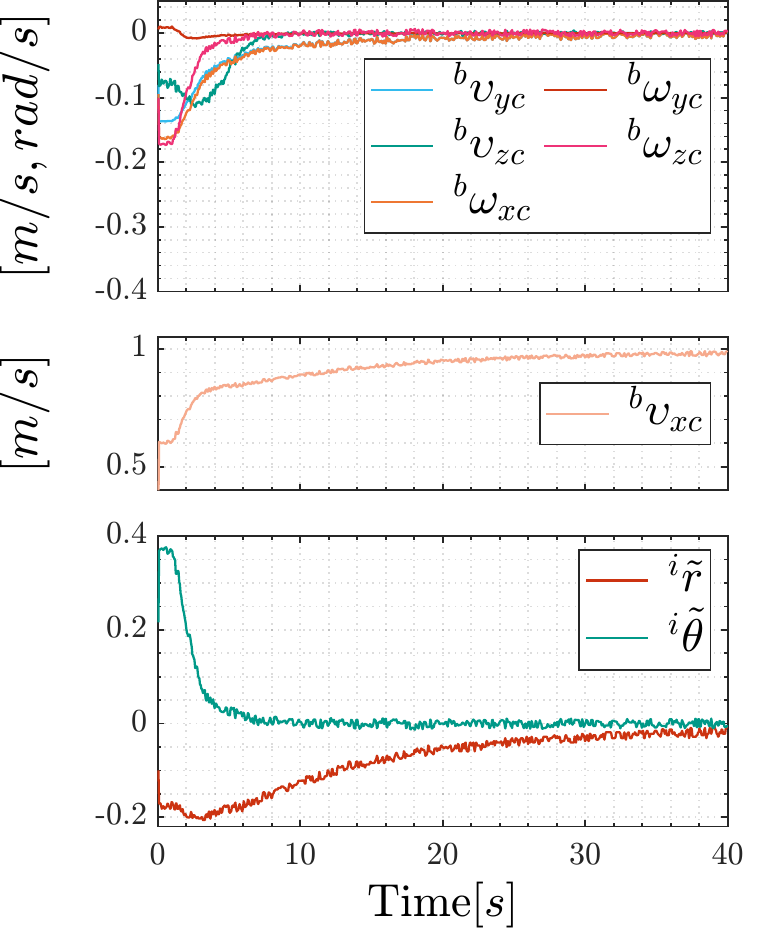}}    
    \hfill
    \centering
    \subfloat[Appropriate selection of matrix $\mathbf{W}$\label{fig:simulation_experiments_findW_b}]{%
        \includegraphics[width=0.4725\linewidth]{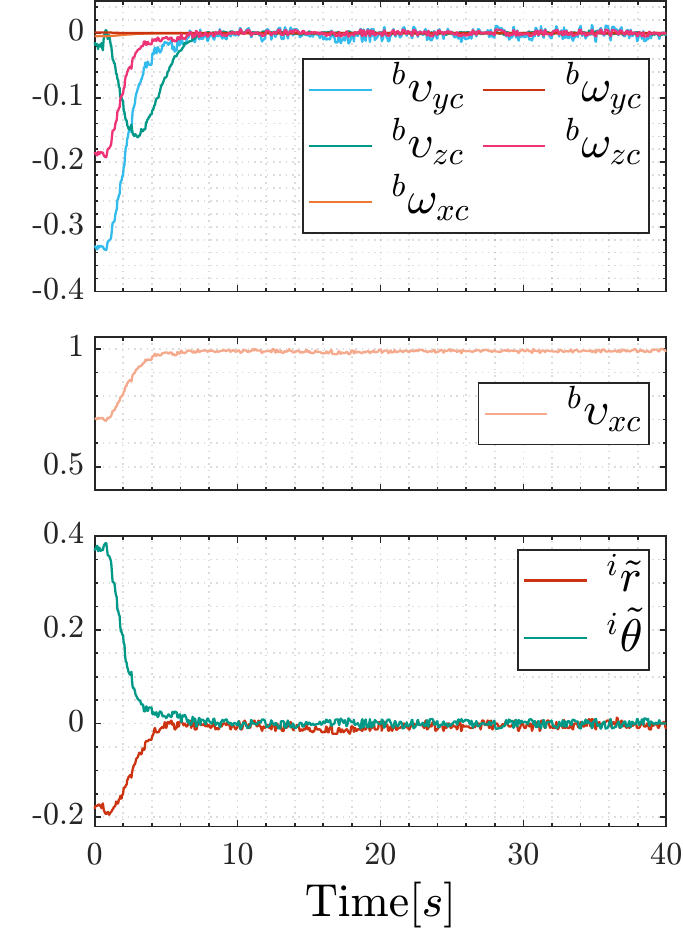}}
    \caption{Simulation results of the visual servoing controller with different values of matrix $\mathbf{W}$.}
    \label{fig:simulation_experiments_findW}
\end{figure}

An inappropriate matrix $\mathbf{W}$ generates control values that do not guarantee $^{b}\omega_{xd} \approx 0$ and $^{b}\omega_{yd} \approx 0$, rendering the visual servoing proposal unsuitable for the aerial platform (DJI Matrice 100). Furthermore, the movement of the aerial vehicle generated by the control law ensures that the error converges to zero $\mbox{ }^{i}\Tilde{\bm{\xi}} = \begin{bmatrix} ^{i}\Tilde{r} & ^{i}\Tilde{\theta} \end{bmatrix} \rightarrow \mathbf{0}$ as $t \rightarrow \infty$, with a stabilization time of $30~[s]$. Finally, the null space projector ensures $^{b}{\upsilon}_{x} \rightarrow~^{b}{\upsilon}_{xd}$ as $t \rightarrow \infty$, with a stabilization time of $30~[s]$ (see results in Fig. \ref{fig:simulation_experiments_findW_a}). 

\begin{figure}[ht]
    \centering
\includegraphics[width=1.0\columnwidth]{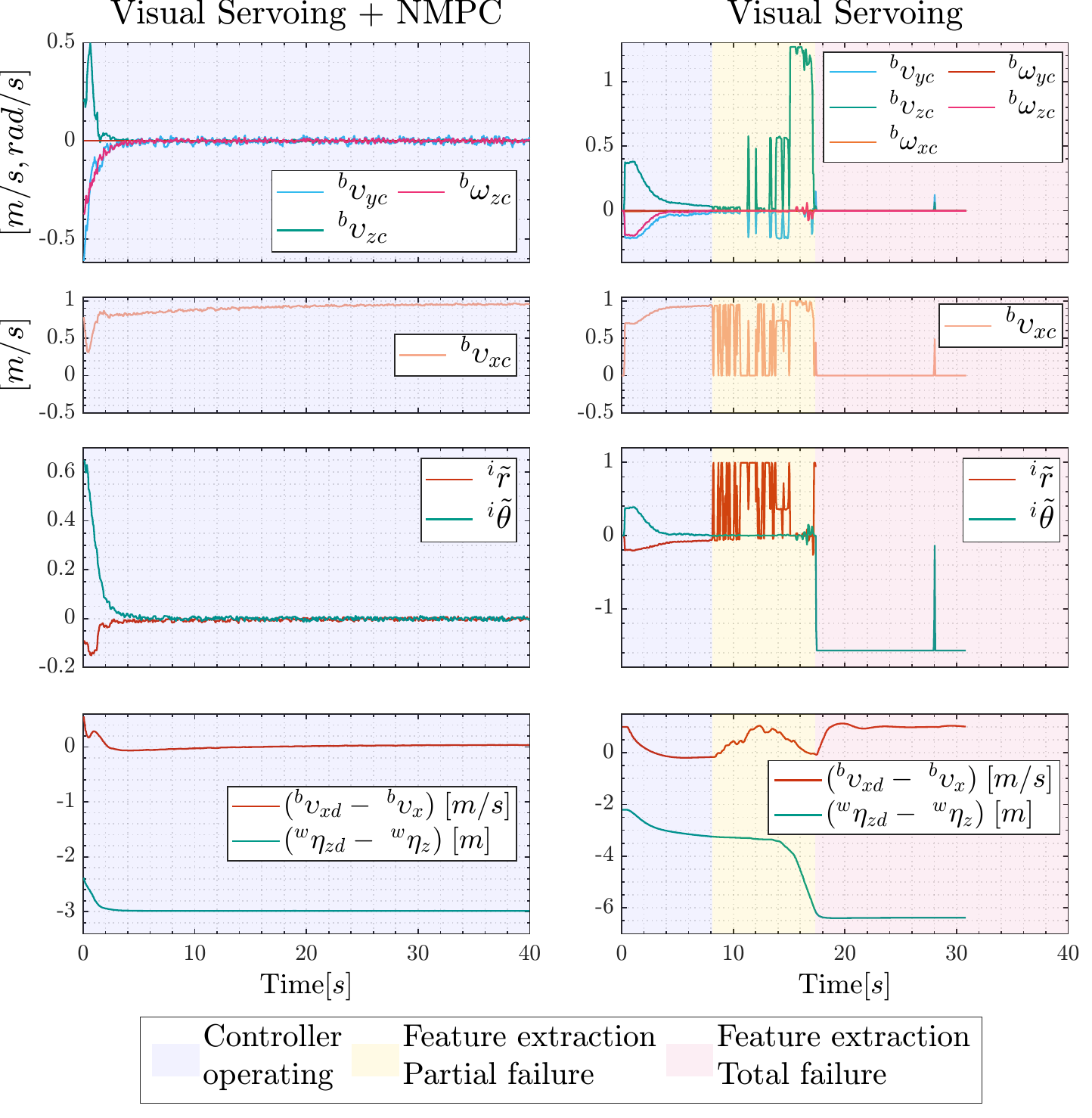}
    \caption{Comparative results of the visual servoing controller versus the visual servoing controller combined with NMPC in a simulated environment.}
    \label{fig:simulation_NMPC_VS}
\end{figure}

On the other hand, an appropriate matrix $\mathbf{W}$ can notoriously improve the controller performance. These values ensure $^{b}\omega_{xd} \approx 0$ and $^{b}\omega_{yd} \approx 0$, guaranteeing that the control error vector converges to zero $\mbox{ }^{i}\Tilde{\bm{\xi}} = \begin{bmatrix} ^{i}\Tilde{r} & ^{i}\Tilde{\theta} \end{bmatrix} \rightarrow \mathbf{0}$ as $t \rightarrow \infty$, with a stabilization time of $7~[s]$. Finally, the null space projector also ensures that $^{b}\upsilon_{x} \rightarrow~^{b}\upsilon_{xd}$  as $t \rightarrow \infty$ with a stabilization time of $7~[s]$ (see results in Fig. \ref{fig:simulation_experiments_findW_b}).

The second experiment, shown in Fig. \ref{fig:simulation_NMPC_VS}, compares visual servoing \eqref{eq:camera:6} with the visual servoing controller combined with NMPC \eqref{eq:mpc:5}, considering the following values in the null space operator: ${^{b}\upsilon}^{max}_{x} = 1~[m/s]$ and $^w\eta_{zd} = 5[m]$. The limitations of the classical visual servoing controller restrict its ability to define the UAV's workspace. For instance, when a loss of visual features occurs due to a flight altitude outside the workspace, the camera loses line of sight with the visual patterns, decreasing the image feature extraction, and causing controller failures. However, due to the constraints of the NMPC, the altitude of the UAV is limited to $4.5~[m]$ above the photovoltaic array (see Table \ref{table:2}), the control proposal ensures proper extraction and detailed images, guaranteeing that $\mbox{ }^{i}\Tilde{\bm{\xi}} = \begin{bmatrix} ^{i}\Tilde{r} & ^{i}\Tilde{\theta} \end{bmatrix} \rightarrow \mathbf{0}$ and $^{b}\upsilon_{x} \rightarrow~^{b}\upsilon_{xd}$  as $t \rightarrow \infty$, with $^w\eta_z$ within the constraints of the workspace in the NMPC.

In Fig. \ref{fig:simulation_NMPC_VS} the yellow background presents failures in feature extraction, which cause problems in the visual servoing controller; the pink background represents a total failure of image feature extraction when the controller is not working. Finally, the purple background represents the time when the controller is operating correctly with correct image feature extraction, where visual servoing combined with the NMPC exhibits proper behavior during the whole experiment.

\subsection{Real-world experiment results}
\label{subsec:Real-world experiment results}
Several experiments were conducted to demonstrate the performance of both proposal at different heights over PV arrays. The null space operators, ${^{b}\upsilon}^{max}_{x} = 0.5~[m/s]$ and $^w\eta_{zd} = 5~[m]$, over the PV arrays are the same as those presented in the second simulation experiments.
\begin{figure}[ht]
    \centering
\includegraphics[width=1.0\columnwidth]{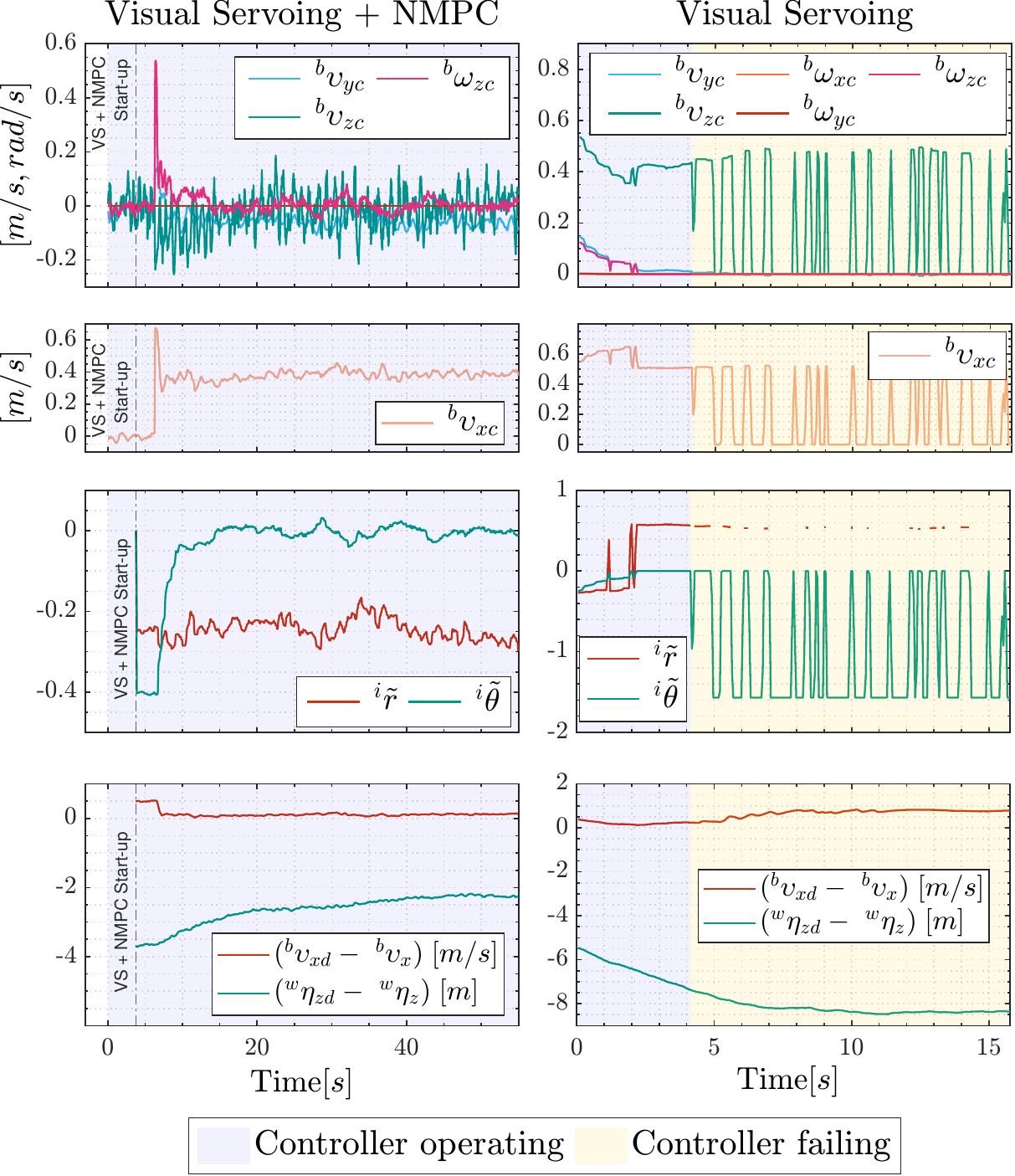}
    \caption{Comparative results of the  visual servoing controller versus the visual servoing controller combined with NMPC in real-world environments.}
    \label{fig:real_experiments}
\end{figure}

Due to the fact that the visual servoing controller does not include the dynamics of the aerial vehicle and the constraints where the image extraction features work properly, this formulation exhibits poor behavior resulting in the controller failure during the experiment (see Fig. \ref{fig:real_experiments}).

On the other hand, visual servoing combined with the NMPC formulation exhibits proper behavior of the aerial vehicle, (see Fig. \ref{fig:real_experiments}), guaranteeing movement along the PV arrays in the real-world application. However, the control error vector exhibits bounded values, with $|^{i}\Tilde{r}| < 0.3 $ and $ |^{i}\Tilde{\theta}| < 0.08$. This behavior is attributed to potential unmodeled dynamics, disturbances during the experiment, the precision of depth camera measurements at this height, and incorrect gain tuning. Additionally, the proposal is able to guarantee $^{b}{\upsilon}_{x} \rightarrow~^{b}{\upsilon}_{xd}$ with $^w\eta_z$ inside the workspace. 

\begin{figure}[ht]
    \centering
\includegraphics[width=1.0\columnwidth]{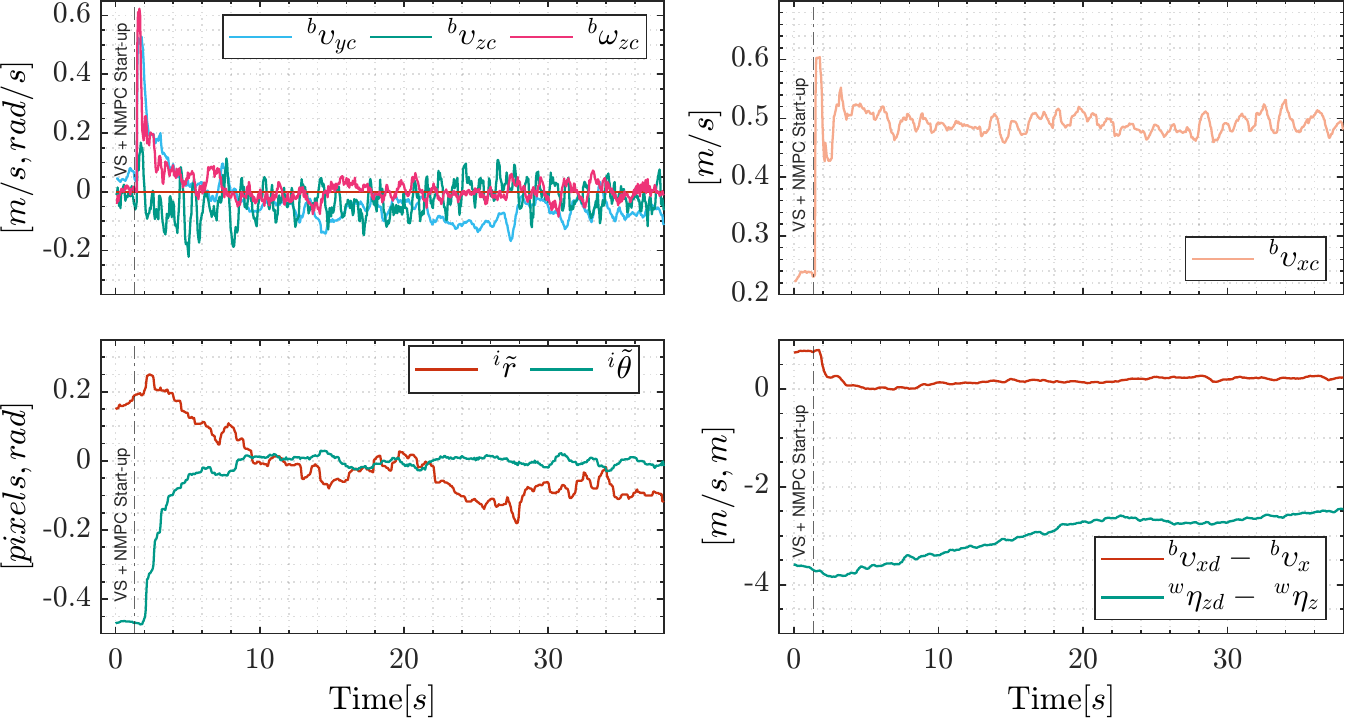}
    \caption{Results of visual servoing controller combined with NMPC with appropriate gain $\mathbf{K}$ to decrease the $^{i}\Tilde{r}$ translation error.}
    \label{fig:NMPC_final_experiment}
\end{figure}

Given the performance demonstrated in the experimental results presented above, the authors adjusted the gain values, setting $\mathbf{K} = \text{diag}(300, 0.8)$. Simultaneously, modifications were applied to the null space operator, with \hbox{${^{b}\upsilon}^{max}_{x} = 0.5~[m/s]$} and $^w\eta_{zd} = 3~[m]$. We have chosen to set the inspection height at $3~[m]$ as it is deemed adequate. This height ensures the minimum distance within the camera's field of view while maintaining a safe flying altitude. Following these changes, controller performance showed a significant improvement, as indicated in Fig. \ref{fig:NMPC_final_experiment}. This achievement is reflected in a maximum final feature error of $|^{i}\Tilde{r}| < 0.18 $ and $ |^{i}\Tilde{\theta}| < 0.05$ that can also guarantee $^{b}{\upsilon}_{x} \rightarrow~^{b}{\upsilon}_{xd}$.

\begin{figure}[ht]
    \centering
\includegraphics[width=1.0\columnwidth,]{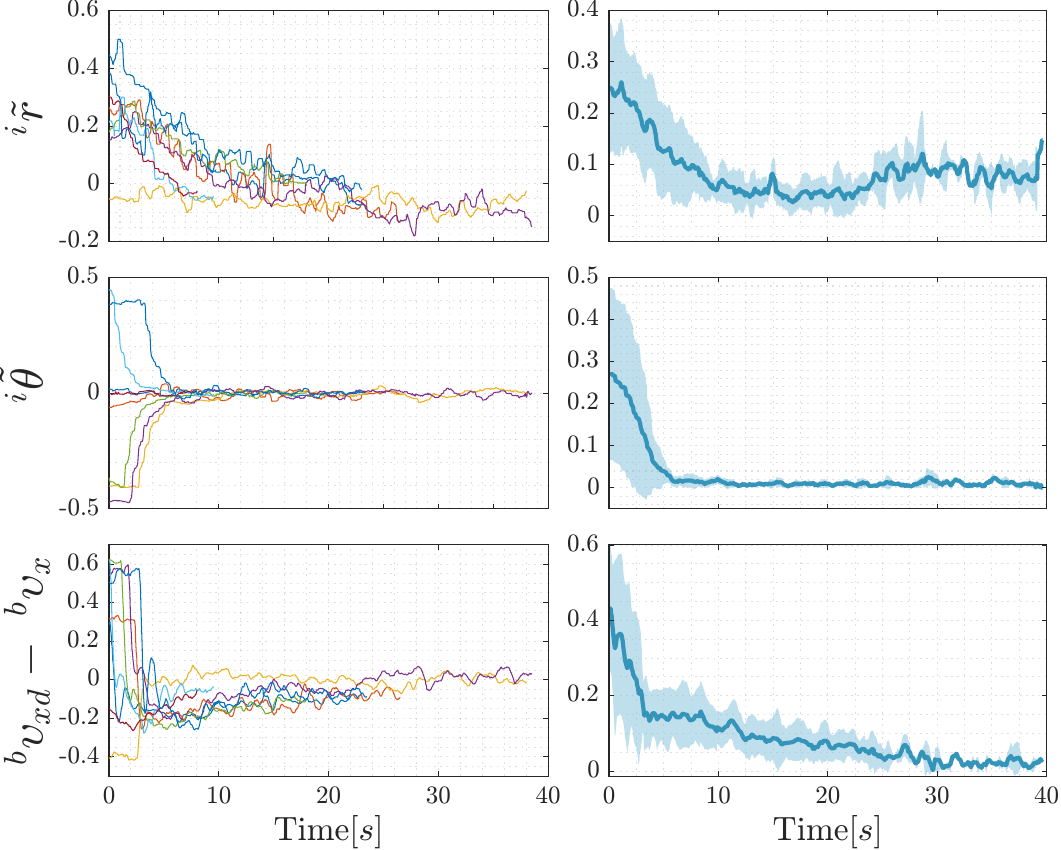}
    \caption{$^{i}\Tilde{r}$, $^{i}\Tilde{\theta}$ and $^b{\Tilde{\upsilon}_{x}}$ errors of eight experiments of the visual servoing controller combined with the NMPC controller in real world environments. The right plots show the mean and standard deviation of the experiments.}
    \label{fig:real_eigth_experiments}
\end{figure}

The improvement in performance is directly associated with changes in the gain matrix and the height of the PV arrays, both influenced by the unmodeled dynamics of the system and the measurements of the depth camera. Fig \ref{fig:real_eigth_experiments} shows the errors $\mbox{ }^{i}\Tilde{\bm{\xi}}$ and $^b{\Tilde{\upsilon}_{x}}$ of eight experiments with different initial conditions using the visual servoing controller with NMPC, where the error vector $[^{i}\Tilde{r}$, $^{i}\Tilde{\theta}]$ is  ultimately bounded, converging to final values close to zero. The right plots in \mbox{Fig \ref{fig:real_eigth_experiments}} show the mean ($\bar{x} = (1/n) \Sigma_{i}^{n}\left| x_i \right|$) and standard deviation ($\sigma^{2}  = (1/(n-1)) \Sigma_{i}^{n}\left | x_i - \bar{x} \right |^{2}$) of the experiments, where these values tend to converge to a value close to zero along the time. These errors are in an acceptable range for the application, thus, the proposal manages to generate images of the PV arrays for an expert inspection process. Further details of the project are presented in the following videos\footnote{\href{https://drive.google.com/file/d/1f1eLxRuRywHLqyqgCv5UXXQRtF1UR9WA/view?usp=sharing}{\scriptsize https://drive.google.com/file/d/1f1eLxRuRywHLqyqgCv5UXXQRtF1UR9WA/view?\\\hspace*{1.5em}usp=sharing}}.

\subsection{Run-time}
Table \ref{tab:run-time} shows the execution time of the approach. The feature extraction, visual servoing controller, and NMPC processes have a total time of $18.1~[ms]$.  Therefore,  we consider that the application runs in real time because the odometry runs at $20~[ms]$ and the camera at $30$ \textit{fps}.

\begin{table} [h]
\centering
\caption{Total run-time of the process in each iteration}
\renewcommand{\arraystretch}{1.2}
\begin{tabular}{l c} 
\hline
\hline
\textbf{Process} & {\textbf{Average time (ms)}} \\
\hline
Feature Extraction: & 13.9 \\
Visual Servoing Controller: & 1.2 \\
NMPC: & 3.0 \\
\hline
\textbf{Total} & 18.1 \\
\hline
\end{tabular}
\label{tab:run-time}
\end{table}
%%%%%%%%%%%%%%%%%%%%%%%%%%%%%%%%%%%%%%%%%%%%
\section{Conclusions}
\label{sec:Conclusions}
We introduce a visual servoing controller integrated with NMPC for the Matrice 100 UAV's overflight of a PV array. The combination of controllers allows the UAV to have a defined workspace; in this way, the UAV mounted camera has a direct line of sight to the PV array allowing a correct extraction of features for the visual servoing controller. In addition, using a Kalman filter aids in estimating the filtered visual features of RGB-D images, specifically the vertical edges of PV solar panels, facilitating the UAV's stable positioning over the array. This correct positioning enables low-altitude flights for capturing detailed images crucial for expert or AI-based identification of potential panel failures.

In future work, we will improve the detection of PV array vertical lateral edges using neural networks on RGB images. This could decrease the processing run-time and noise in the extraction of visual features. In addition, we will investigate the integration of the vision system and the dynamics of the UAV into a single optimization problem (Perception NMPC).

\section*{Acknowledgements}
This work has been supported by the Ministry of Science and Innovation of the Spanish Government through the research project PID2021-122685OBI00 and the grant PRE2019-088069 for training PhD research staff. 

In addition, we are grateful for the support of the Secretariat of Science, Technology and Innovation (SECITI) - San Juan, Argentina (Project: PIPE 2022) and the staff of the Anchipurac Photovoltaic Solar Park (EPSE) located in the same city, who provided us with the facilities to perform the flight tests with the UAV.  

The results of this work are part of the project “Tecnologías de la Industria 4.0 en Educación, Salud, Empresa e Industria” developed by Universidad Tecnológica Indoamérica.
%%%%%%%%%%%%%%%%%%%%%%%%%%%%%%%%%%%%%%%%%%%%%%%

\bibliography{references.bib}{}
\bibliographystyle{IEEEtran}
\end{document}